# Automated Characterization of Stenosis in Invasive Coronary Angiography Images with Convolutional Neural Networks

*Benjamin Au MS[1], Uri Shaham PhD[1], Sanket Dhruva MD[1], Georgios Bouras PhD[2], Ecaterina Cristea MD[2], Alexandra Lansky MD PhD[2], Andreas Coppi PhD[1], Fred Warner PhD[1], Shu-Xia Li PhD[1], Harlan Krumholz MD SM[1]*

## Abstract

The determination of a coronary stenosis and its severity in current clinical workflow is typically accomplished manually via physician visual assessment (PVA) during invasive coronary angiography. While PVA has shown large inter-rater variability, the more reliable and accurate alternative of Quantitative Coronary Angiography (QCA) is challenging to perform in real-time due to the busy workflow in cardiac catheterization laboratories. We propose a deep learning approach based on Convolutional Neural Networks (CNN) that automatically characterizes and analyzes coronary stenoses in real-time by automating clinical tasks performed during QCA. Our deep learning methods for localization, segmentation and classification of stenosis in still-frame invasive coronary angiography (ICA) images of the right coronary artery (RCA) achieve performance of 72.7% localization accuracy, 0.704 dice coefficient and 0.825 C-statistic in each respective task. Integrated in an end-to-end approach, our model's performance shows statistically significant improvement in false discovery rate over the current standard in real-time clinical stenosis assessment, PVA. To the best of the authors' knowledge, this is the first time an automated machine learning system has been developed that can implement tasks performed in QCA, and the first time an automated machine learning system has demonstrated significant improvement over the current clinical standard for rapid RCA stenosis analysis.

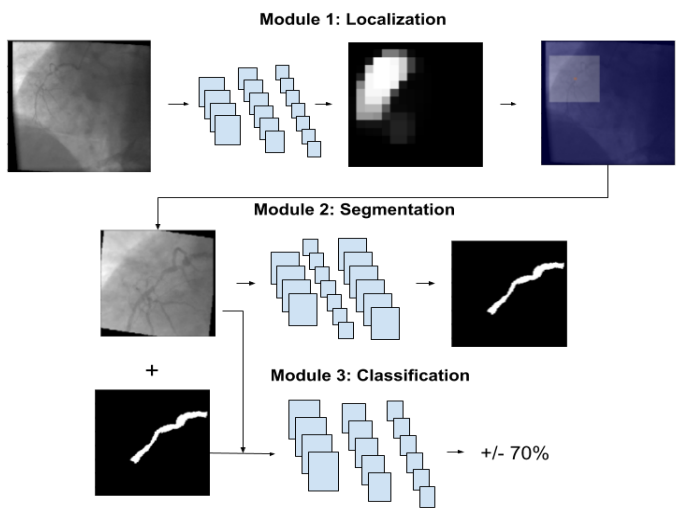

**Figure 1. Overview of Experimental Method.** Module 1: Localize a 192x192 bounding box around the stenosis lesion in the 512x512 image. Module 2: Segment the diseased lesion in the localized image. Module 3: Classify the combined image and segmentation input binarily via 70% stenosis threshold.

## 1. Introduction

Coronary artery disease (CAD), a leading cause of death worldwide and a target for coronary revascularization, involves narrowing (i.e. stenosis) of at least one of the large arteries supplying the heart. One method of determining the presence and severity of coronary stenosis is invasive coronary angiography (ICA), which involves continuous X-ray (i.e. fluoroscopy) with simultaneous injection of radiopaque contrast into the coronary arteries. A diagnosis of stenosis severity of ≥50% by diameter is generally considered the minimum standard for considering the decision to pursue coronary revascularization of the three large epicardial arteries, left anterior descending (LAD), left circumflex (LCx) and right coronary arteries (RCA), when accompanied by functional assessment. Without functional assessment, ≥70% diameter stenosis is generally considered an indicator in identifying clinically significant lesions [1]. Stenosis severity is typically determined in clinical practice by physician visual assessment (PVA). This approach, while considered as clinical standard, has known limitations, such as significant intra- and inter-rater variability, as well as high positive prediction bias, implicitly leading to overutilization of clinical services [2]. A more reliable and accurate determination of stenosis severity is quantitative coronary angiography (QCA) [2,3].

*1, Yale Center for Outcomes Resesarch and Evaluation; 2. Yale Cardiovascular Research Group*

QCA is a computer-assisted procedure widely considered as a gold standard for measuring stenosis percentage. It is typically performed either 'online' immediately after coronary angiography for clinical decision making or 'offline' by angiographic core laboratory experts (for clinical trials) and involves visually annotating diseased coronary arterial segments and the area surrounding each stenosis to determine the percent diameter stenosis. Standard workflow for QCA consists of a multi-step analytical pipeline. Single-frame images that best demonstrate the stenosis are selected by the analyst. Images are then manually annotated by segmenting a portion of the lesion (i.e. stenosis) of interest, consisting of the lesion as well as the surrounding healthy vessel. Finally, annotated lesions are analyzed for percent diameter stenosis relative to the reference vessel diameter of the lesion of interest [4,5].

While the QCA measurement process involves certain subjective decision-making components and may be challenging in measuring complex lesions (such as with thrombus and calcification), it currently offers the most accurate and reproducible measurements of anatomical coronary stenosis severity and thus is considered the clinical gold standard for measuring coronary stenosis. However, despite these improvements over PVA, QCA has not been widely implemented into the busy workflow of cardiac catheterization laboratories, due to it being a manual, human resource-intensive procedure [3].

These clinical challenges motivate the development of automated tools for rapidly, accurately and reliably analyzing coronary stenoses. In recent years, Convolutional neural networks (CNN) methods have demonstrated highly accurate and reliable performance across a variety of computer-vision related tasks, including image classification [6–8], object detection [9–11], and semantic segmentation [12–14]. In spite of their great success in other imaging domains, gold-standard CNN's such as VGG16 [6] and ResNet50 [8] simply fail to learn 'off-the-shelf' when attempting to directly classify stenosis from raw ICA images, according to baseline experiments. This is influenced in part by the technically challenging aspects of ICA imaging: ICA images contain significant image noise, at time including non-anatomical artifacts such defibrillator wires, and have limited image contrast. Furthermore, diseased lesions in ICA images represent a fraction (~0.25% in our dataset) of the overall image, adding a non-trivial search component to the image analysis problem.

Motivated by the opportunity to improve clinical care by offering a more accurate and rapid way to quantify coronary stenoses, we apply deep learning CNN techniques to develop improved methods for characterizing stenosis by replicating the QCA analytical process in a three-step analytical pipeline. Our methods seek to automate key tasks of the QCA workflow, solve the issues of poor inter-rater agreement, and can be run in real-time, all while providing high-quality annotations and stenosis characterization at each step (see Figure 1). To the authors' knowledge, this is the first attempt to automatically characterize and analyze the intensity of stenosis from invasive coronary angiography with deep learning methods.

Our contributions are as follows: We propose three CNN models to solve the problems of localization, segmentation and classification with strong performance. We then integrate these models and develop an end-to-end deep learning pipeline for characterizing stenosis in right coronary artery ICA images. Finally, we compare model performance relative to the current clinical standard of PVA, using performance measures reported in [2].

## 2. Related Work

CNN methods have been widely applied to medical imaging in cardiology across several imaging modalities, including left ventricle and structure segmentation in MRI [15–20], heart chamber segmentation [21], calcium scoring [22], and stenosis detection [23] in CT imaging, as well as vessel extraction [24] in invasive coronary angiography. Much of previous work using deep learning methods applied to medical imaging in cardiology is focused on segmentation-related tasks.

The authors of [23] focus on the clinical problem of identifying functionally relevant stenoses in the coronary vasculature, and also utilize deep learning methods, e.g. CNNs, to build a localization-segmentation-classification analytical framework similar to ours, all while obtaining promising results. However, there are a few key differences between our respective work. The authors in [23] characterize the left ventricle to obtain functional data, while we focus directly on the right coronary artery to obtain anatomic measurements. Furthermore, [23] utilize image volumes captured from Coronary CT angiography (CCTA), while we use data from invasive coronary angiography. CCTA images can be viewed as three dimensional volumes, while invasive coronary angiography images are 2D, thus motivating separate deep learning methods for image analysis. In particular, localization-segmentation subtasks are performed in [23] by pixel-wise classification via three CNN's, analyzing image slice contexts across each of the three axial views in tandem, whereas we perform segmentation and localization tasks using separate CNN's in a single pass. Finally, invasive coronary angiography and CCTA both serve distinct and important roles in the clinical setting. Invasive coronary angiography allows patients to receive percutaneous coronary intervention (PCI), whereas CCTA is a non-invasive imaging modality to evaluate patients for coronary artery disease. In summary, the work in [23], while similar, is distinct and complementary in imaging modality, deep learning methodology, and clinical applicability, relative to our work.

Automatic detection in coronary stenosis has also been performed through other methods in invasive coronary angiography such as deformable splines [25] and spatio-temporal tracking [26]. Specifically, the work in [26] develops end-to-end characterization of the presence of stenosis in right coronary artery with high sensitivity (90%) and specificity (86%). However, [26] develops methods to distinguish stenosis from non-stenosis, whereas our goal is to distinguish anatomically relevant stenosis for PCI from non-anatomically relevant stenosis. The latter is more clinically relevant in cardiology because of its relevance to determining need for revascularization. In addition, the algorithms in [26] do not learn from data, and algorithmic performance on unseen test data is not stated.

### 3. Cohort

The dataset for this work consists of ICA imaging data of the RCA of 1024 study participants. Participant data were retrospectively aggregated by the Yale Cardiovascular Research Group, and de-identified and obtained for use with IRB Exemption without need for participant consent. Subjects were selected from a range of prospective trials evaluating interventional coronary devices in adult patients aged 18 and older, with stable coronary disease or acute coronary syndromes (ST Elevation Myocardial Infarction were excluded). All angiograms were acquired with a prespecified acquisition protocol for optimal angiographic quality.

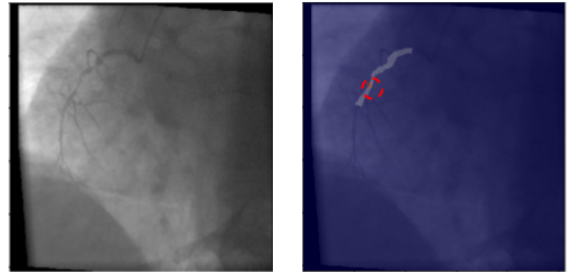

**Figure 2. ICA image with stenosis annotations.** Left: Invasive coronary angiography image with data augmentation. Right: Image with lesion of interest mask (white) and stenosis coordinate (encircled in red) overlaid. Stenosis for this example is 78%.

The data for each participant consists of the following: 1) de-identified single-frame image extracted from invasive coronary angiography 2) lesion segmentation annotations (LSA) corresponding to each image 3) coordinate location of stenosis and 4) QCA stenosis measurement, 5) normalized lesion silhouette annotation (see Figure 5). For a visual example of data, see Figure 2. Across the full dataset, the basic statistical properties for the set of stenosis measurements are as follows: mean: 74.3%, standard deviation: 18.8%, min: 4.3%, max: 100%.

### 3.1 Preprocessing

We preprocess the dataset in a few steps to standardize certain heterogeneous data components. First, images that contained more than one annotated stenosis lesion or no annotated stenosis lesion were excluded. This exclusion criterion retained 90% of the original dataset. Second, angiography images and segmentation annotations were contrast standardized, i.e. minimum and maximum pixel range were set to 0 and 1, respectively. Finally, image sizes were standardized to $512 \times 512$ pixels.

### 4. Methods

To identify participants with anatomically significant stenosis, we first localize the invasive coronary angiography image by learning to place a bounding box around the stenosis and consequently reduce the size of the image. We then perform segmentation on the localized image. Finally, we classify the stenosis based on the combined invasive coronary angiography image and segmentation annotation, both of which are localized. Each task is performed by a separate CNN. The pipeline is visualized in Figure 1.

## 4.1 Localization

We build a CNN for the localization task inspired YOLONet [11], which has demonstrated near state-of-the-art real-time object detection and localization on the PASCAL VOC 2007 dataset [27]. The CNN contains 22 convolutional layers and 5 max pooling layers, along with residual skip connections. We simplify the CNN output layer so that for each image input, the network produces a $k \times k$ array of confidence scores $O = \{o_{ij} : i,j = 1, ..., k\}$ ; $0 \leq o_{ij} \leq 1$, which are used to predict the presence of stenosis within $64 \times 64$ regions of the image, tiled with a 32 pixel stride in each direction. The net is trained on weighted sigmoid cross-entropy loss on each cell output logit

$$loss(O, Y) = - \sum_{i,j=1,...,k} [w * y_{ij} log(\sigma(o_{ij})) + (1 - y_{ij}) log(1 - \sigma(o_{ij}))] \quad (1)$$

where $\sigma(x) = \frac{1}{1+e^{-x}}$ is the sigmoid function and $y_{ij} = 1$ if the $(i,j)$ region contains the stenosis and $y_{ij} = 0$ otherwise. The positive weight coefficient $w$ is defined as $k^2$ to offset class imbalance in each image.

Once the CNN is trained, we obtain a $192 \times 192$ pixel bounding box prediction using the following procedure: The output grid is transformed into a non-overlapping $16 \times 16$ grid, where each cell is associated with a score $p_{ij} = max(o_{ij}, o_{i(j-1)}, o_{(i-1)j}, o_{(i-1)(j-1)})$. Then, the $192 \times 192$ bounding box is selected by choosing the region with maximum sum component score $R = argmax_{i,j} [\sum_{m=i}^{i+5} \sum_{n=j}^{j+5} p_{mn}]$.

Visualization of grid and final localization can be seen in Figure 5.

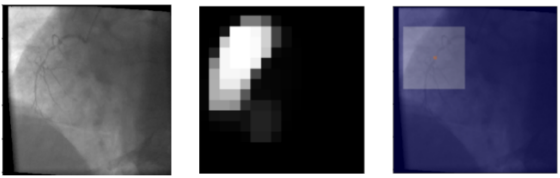

**Figure 3. Visualization of Localization Process.** Left: input invasive coronary angiography image. Middle: $16 \times 16$ non-overlapping grid obtained from CNN output. Right: Image with 192x192 bounding box and stenosis coordinates overlaid.

## 4.2 Segmentation

For the segmentation task, we build a CNN based on U-net [13], motivated by the model's state-of-the-art performance in other biomedical segmentation tasks. U-net is structured as an encoder, which consists of five convolutional blocks, followed by a decoder, which consists of five deconvolutional blocks, with skip connections between the respective encoder-decoder blocks. In our proposed approach, we use average pooling instead of max pooling, which provides a slight performance boost. The network uses single channel $192 \times 192$ localized input and outputs a binary $192 \times 192$ array $U = \{u_{ij} : i,j = 1, ..., 192\}$. We train the net using dice coefficient loss (see Equation 2), which is a measure of segmentation fidelity and has previously been used in other biomedical image segmentation tasks such as [28]

$$dice\_loss(U, V) = \frac{-2 \sum_{i,j=1,...,192} [u_{ij} * v_{ij}]}{\sum_{i,j=1,...,192} [u_{ij}] + \sum_{i,j=1,...,192} [v_{ij}]} \quad (2)$$

where $V$ corresponds to the ground truth obtained from the lesion segmentation annotations from the QCA. A visualization of the input, output and predicted segmentation of the segmentation CNN is shown in Figure 4.

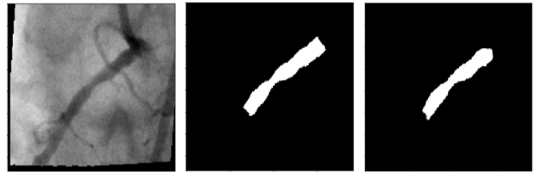

**Figure 4. Visualization of Segmentation.** Left: Input localized ICA image. Middle: Ground truth stenosis lesion annotation. Right: Predicted annotation.

In addition, we develop a morphologically-weighted sigmoid cross entropy (MWCE) loss designed to generate sharp edge segmentation predictions around the diseased lesion. The loss function is designed follows. We generate $k \times k$ array of confidence scores $O = \{o_{ij} : i,j = 1, ..., k\}$ ; $0 \leq o_{ij} \leq 1$, where $k$ is the number of pixels in a single image axis (in this case 192), representing a prediction of the presence of the lesion in the given pixel. The net is trained on a MWCE loss summed over each pixel output logit (see Equation 3).

$$MWCE = - \sum_{i,j=1,...,k} w_{ij} [y_{ij} log(\sigma(o_{ij})) + (1 - y_{ij}) log(1 - \sigma(o_{ij}))] \quad (3)$$

$$w_{ij} = \alpha o_{ij} + b_{ij} m_{ij} \quad (4)$$

$$b_{ij} = \beta y_{ij} + \gamma z_{ij} + (1 - y_{ij} - z_{ij}) \quad (5)$$

$$m_{ij} = 1 + \delta * exp(-\frac{(c_0-i)^2 + (c_1-j)^2}{\sigma^2}) \quad (6)$$

In words, the morphological weighting is a pixel-wise weighting scheme that promotes precise segmentation of stenosis lesion, particularly near the point of stenosis, while regularizing for false positives. It consists of false positive regularization term $\alpha o_{ij}$, (where $\alpha$ is a hyperparameter) as well as the core weighting component $b_{ij} m_{ij}$ that promotes accurate segmentation of the lesion and its silhouette near the point of stenosis. This latter component consists of a base subcomponent $b_{ij}$ that is weighted more heavily if the pixel is part of the lesion segment $y_{ij}$ ($y_{ij} = 1$ if the $(i,j)$ pixel is part of the lesion and $y_{ij} = 0$ otherwise), as well as if the pixel is part of the normalized silhouette surrounding the lesion $z_{ij}$ ($z_{ij} = 1$ if the $(i,j)$ pixel is part of the normalized silhouette (see Figure 5) and $z_{ij} = 0$ otherwise), with large hyperparameter values chosen for $\beta, \gamma$ (see Equation 5). For visualization of the normalized silhouette annotation, see Figure 5. The multiplier subcomponent $m_{ij}$ further promotes accurate segmentation near the point of stenosis by augmenting the base subcomponent weighting with a Gaussian multiplier centered around the point of stenosis $(c_0, c_1)$, with hyperparameters for peak intensity $\delta$ and bell width $\sigma$ (see Equation 6).

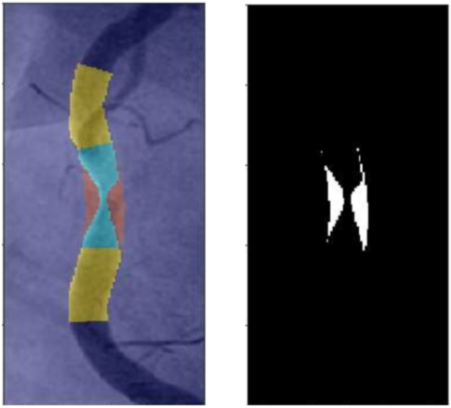

**Figure 5. Visualization of stenosis silhouette component.** Left: The standard lesion segmentation annotation (LSA) in segmentation consists of diseased and healthy components, colored in sky blue and tan respectively. Right: Normalized Silhouette Annotation. This corresponds to the rose-colored region in the left image. We use this for calculating morphologically-weighted sigmoid cross entropy loss (see Equation 3).

### 4.3 Classification

For the classification task, we train a small CNN, containing five convolutional layers, and a single global max pooling layer with a single fully-connected layer prior to obtaining output. Image features are globally pooled prior to being connected to a final fully-connected layer. See Figure 5 for a visualization of the model architecture. Training is performed using mean squared error regression on stenosis percentage, which is a richer data format relative to binary classification at the 70% stenosis threshold indicator for revascularization.

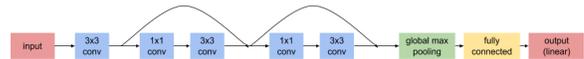

**Figure 5. Diagram of Proposed CNN Architecture for Classification Task.** Skip connections are concatenated as in DenseNet [29]. Network is trained via regression on ordinal stenosis values and evaluated via binary classification at the 70% stenosis threshold.

### 4.4 Data Augmentation

Data augmentation techniques are applied during the training phase of each CNN. The techniques include random shift, random rotation, random shear (see Appendix A for details). Random flip and elastic deformation augmentation techniques were considered and ultimately not used as they were non-improvements in training speed and performance. Flips were excluded due as flipped images are never observed due to image acquisition protocol. Elastic deformation were excluded, since artery warping directly impacts the degree of stenosis.

### 5. Results

In the localization task, we predict the location of a pixel subregion from the original image. This subregion size is designed to be small enough to gain performance on segmentation and large enough to gain performance on classification. In the segmentation task, we train a CNN to generate a high-fidelity stenosis lesion annotation, as is done during QCA, from a localized ICA image input. In the classification task, we classify the stenosis based on whether the stenosis is ≥70%, which is considered an indicator for clinical validation in determining need for revascularization. Here, we use the localized ICA image and lesion annotation as inputs. Finally, we run the deep learning pipeline end-to-end, training a network to classify stenosis solely on ICA images while producing stenosis lesion annotations. We split our dataset via stratified random sampling based on stenosis degree into 70:15:15, train, validation, and test sets. Hyperparameters are explored with the training and validation sets and final performance is reported from a separate test set (see Appendix A for technical details and hyperparameter specification).

## 5.1 Evaluation

We evaluate the localization task with localization accuracy, defined as the frequency at which the proposed localization region contains the stenosis. We evaluate the segmentation task using dice coefficient, which measures general fidelity of segmentation. We measure the classification and end-to-end task using C-statistic, i.e. area under the Receiver Operating Characteristic curve, by binarizing the stenosis at the 70% threshold. Additional metrics for comparing our end-to-end model to PVA as reported in [2] are false discovery rate, defined by $\frac{FalsePos}{FalsePos + TruePos}$, and assessment bias, i.e. the difference between ground truth and prediction. Standard deviation measures are obtained by evaluating model performance across 5 runs with distinct random seeds.

## 5.2 Localization

| Localization Accuracy | | |
|---|---|---|
| **Net** | **Localization Region Size** | |
| | 128x128 | 192x192 |
| **Proposed CNN** | 55.6% (+/- 3.2%) | 72.7% (+/- 2.9%) |
| ResNet50 (Coordinate Regression) | 42.6% (+/- 3.7%) | 66.7% (+/- 3.0%) |

**Table 1: Localization Accuracy Results.** Results show the percentage of times in which the point of stenosis lies within the predicted localization region of the given size. Values are presented as *mean +/- standard deviation*.

For the localization task, we seek to predict a square localization region that contains the point of stenosis. Table 1 shows the localization performance of our proposed approach, where we display the percentage of localization regions that contain the stenosis for two region sizes: $128 \times 128$ and $192 \times 192$. As a baseline comparison, we add corresponding results obtained by a ResNet50 [30], which performs localization by mean squared error regression to explicitly predict coordinates of stenosis. We find that our proposed approach achieves significant improvement in localization over baseline with localization accuracy of 55.% and 72.7% for $128 \times 128$ and $192 \times 192$ bounding box sizes, respectively.

## 5.3 Segmentation

For the segmentation task, we seek to generate high-fidelity segmentation of the lesion of interest given a pre-localized ICA image input. Table 2 shows the segmentation performance of our proposed approach based on [13], which we compare to a fully-convolutional DenseNet (DenseNet FCN) [31], as well as a CNN employing atrous convolutions of multiple rates, DeepLabv3 [32]. Both of these CNN baselines have achieved state-of-the-art in semantic segmentation tasks. We find that our experimental model performs comparable to the DenseNetFCN at 0.704 dice coefficient, and performs significantly better than DeepLabV3.

| Segmentation Results | |
|---|---|
| **Net** | **Dice Coefficient** |
| **Proposed CNN (Dice)** | **0.704 +/- 0.017** |
| **Proposed CNN (MWCE)** | **0.660 +/- 0.026** |
| DenseNet FCN | 0.699 +/- 0.014 |
| DeepLabV3 | 0.657 +/- 0.016 |

**Table 2. Segmentation Results.** Our experimental approach based on U-Net performs comparably to the DenseNet FCN baseline and outperforms the DeepLabV3 baseline, using Dice Coefficient as the evaluation metric. Proposed CNN using MWCE loss performs significantly worse when evaluated using Dice Coefficient, but improves end-to-end performance (see Section 5.5). Values are presented as *mean +/- standard deviation*.

We also run segmentation experiments using the morphologically-weighted cross entropy (MWCE) loss in Equation 3. We find that the Proposed CNN achieves a significantly lower dice coefficient than when optimized using dice loss at 0.660, segmentation using MWCE is significantly improved during end-to-end experiments (see Section 5.5). This is because, MWCE loss is designed to obtain more precise segmentation near the point of stenosis with reduced segmentation accuracy in regions further from stenosis, in contrast to dice loss, which is designed for overall segmentation accuracy. Visual analysis confirms that this is indeed true (see Figure 6).

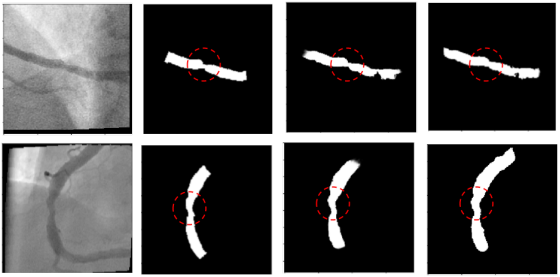

**Figure 6.** Visual Evaluation of Segmentation Using Morphologically Weighted Cross Entropy (MWCE) vs. Dice Loss. Two rows show two example image segmentation instances. Col1 shows input image; Col2 shows ground truth segmentation; Col3 shows segmentation via MWCE loss; Col 4 shows segmentation via dice loss. MWCE predictions shows improved segmentation near the point of stenosis (circled in red), leading to improved ability to measure stenosis intensity.

### 5.4 Classification

| Classification Results | |
|---|---|
| **Net** | **C-Statistic** |
| **Proposed CNN** | **0.825 +/- 0.059** |
| ResNet50 | 0.629 +/- 0.059 |

**Table 3. Classification Results.** C-statistic, i.e. AUROC, is used as evaluation metric. The experimental CNN achieves high classification performance, significantly above the ResNet50 baseline. Both are trained via mean squared error regression. Values are presented as *mean +/- standard deviation*.

For the classification task, we seek to determine whether image inputs meet the clinical criterion for revascularization, i.e. ≥70% stenosis, provided pre-localized invasive coronary angiography images and segmentation annotation two-channel input. Table 3 shows the classification performance of our Proposed CNN, described in Section 4.3, relative to a baseline of ResNet50 [30] utilizing global max pooling of feature outputs with network weights pre-trained on ImageNet. For each CNN model, we train via run regression on ordinal stenosis values and evaluate the model via C-statistic over the binarizing stenosis values at the 70% threshold. The Proposed CNN model significantly outperforms the ResNet50 regression baseline, achieving C-statistic of 0.825.

### 5.5 End-to-End

| End-to-End Results | |
|---|---|
| **Net** | **C-Statistic** |
| **MWCE-End-to-End Pipeline** | **0.578 +/- 0.021** |
| **MWCE-Pre-localized Segmentation-Classification** | **0.703 +/- 0.033** |
| Dice-End-to-End Pipeline | 0.530 +/- 0.021 |
| Dice-Pre-localized Segmentation-Classification | 0.591 +/- 0.045 |

**Table 4. End-to-End Results.** End-to-end pipeline achieves a modest boost above baseline approaches, using C-statistic as an evaluation metric. Each model is trained via mean squared error regression for its classification module, binarizing predicted output at the 70% stenosis threshold for C-statistic evaluation. Dice-models use dice coefficient loss in their segmentation modules, while MWCE-models use morphologically-weighted cross entropy loss (see Equation 3). Values are presented as *mean +/- standard deviation*.

Finally, we combine the individual tasks to form an end-to-end deep learning pipeline (see Figure 1), classifying stenosis based solely on the full still-frame invasive coronary angiography image input. The end-to-end pipeline is pre-trained with weights from the individual tasks, with localization and segmentation modules frozen to prevent overfitting. We run end-to-end experiments using two model types, with Dice-models and MWCE-models, using dice coefficient loss and MWCE loss in their respective segmentation modules. Localization modules output localization regions of size $192 \times 192$, chosen based on overall end-to-end performance. Table 4 shows the classification results of our proposed end-to-end deep learning method.

As mentioned in Section 1, simple baseline CNN's such as ResNet and our Proposed CNN from Section 5.4 fail to train when attempting to classify stenosis directly from the full ICA still-frame image as a single-channel input. In contrast, our end-to-end pipeline model achieve modest performance boost above baseline, with MWCE-End-to-End achieving a C-Statistic of 0.578. We also show results of Segmentation-Classification models using images pre-localized with a localization oracle and show that, as expected, the MWCE-model gains a performance boost from the localization oracle. In addition, the MWCE-model again performs better than its Dice-model counterpart.

We compare the MWCE-End-to-End model to Physician Visual Assessment vis-a-vis metrics of false discovery rate and mean prediction bias reported by [2]. Relative to the current clinical standard for real-time RCA stenosis analysis, our methodology demonstrates a statistically significant improvement of 36.1% in false discovery rate at the 70% stenosis binarization threshold and and a non-significant improvement of -0.1% ins assessment bias (see Table 5).

| **Predicting Stenosis (AI vs. Physician Visual Assessment)** | | |
|---|---|---|
| Evaluator | False Discovery Rate (@70%) | Assessment Bias |
| **End-to-End Deep Learning Pipeline** | **36.8% +/- 2.1%** | **-0.1% +/- 17.0%** |
| Physician Visual Assessment [2] | 50.6% | 16.0% +/- 11.5% |

**Table 5. Comparison between AI and Physician Visual Assessment.** End-to-end pipeline achieves statistically-significant improvement over physician visual assessment, according to false discovery rate at 70% stenosis threshold and non-significant improvement in assessment bias. Physician Visual Assessment metrics are reported by [2]. Values are presented as *mean +/- standard deviation*.

## 6. Discussion

Our experimental approach for the classification task was inspired by the finding that localized segmentation annotations enabled high performance in stenosis classification when included as a second channel to image input. In addition, we found that CNN's with fewer layers performed better than gold standard CNN's architected based on the ImageNet task, such as ResNet. In particular, our final best-in-class classification architecture contained only five total convolutional layers with a single global max-pooling layer, suggesting that stenosis characterization relies mainly on low-level visual features, which may be obfuscated by deeper CNN's with multiple pooling layers. This is reasonable since stenosis is typically determined by analyzing the narrowing of arterial lesions, which can be reduced to analyzing curve-like image features. Additionally, we found that training via mean squared error regression generally captured greater informatic value and led to superior model performance, when compared with training via binary classification using the 70% stenosis threshold value.

The morphologically-weighted pixel-wise cross entropy loss was inspired by the work of [13], which used a pixel-specific weighting scheme in its cross entropy loss. Our weighting scheme was motivated by the realization that dice coefficient did not promote precise segmentation around stenosis lesion edges, potentially causing drops in classification performance downstream, even if dice coefficient loss showed strong segmentation performance in isolation. Thus, we designed a the MWCE pixel-weighting scheme that placed importance on correctly classifying both diseased lesions pixels as well as their silhouette negative counterparts, with particular weighting emphasis on pixels nearest the point of stenosis, and all while discouraging false positive prediction. The result was a slight worsening in segmentation performance as measured by dice coefficient, but significantly improved classification performance in end-to-end modeling.

While anatomic measurement of stenosis severity for a single lesion seems relatively straightforward, i.e. calculation is based on visual narrowing of the lesion, identifying clinically relevant lesions and accurately determining stenosis percentage requires significant clinical sophistication in a number of ways. First, coronary artery reference diameter, both proximally and distally, is an important factor for determining the percentage of stenosis in QCA. In addition, some coronary arteries will have multiple stenoses, often of varying degrees of severity. In some cases, the most diseased lesion by percent diameter stenosis is not always the most clinically significant lesion, as other factors such as functional assessment (through fractional flow reserve), non-invasive imaging, and clinical history can inform the decision to perform revascularization [33]. Functional assessment, in particular, is important for stenoses of 40% and 70% severity, in which American College of Cardiology/American Heart Association Appropriate Use Criteria define clinical significance for stenoses with a fractional flow reserve <0.80 [1]. Identifying clinically significant lesions, as well as correctly measuring percent diameter stenosis and guiding functional assessment then, is not just a vision problem, but one that also requires substantial clinical knowledge and sophisticated clinical validation techniques as well.

Recent research has shown that physician visual assessment (PVA), the current clinical standard for assessing coronary stenosis, does in fact have high intra- and inter-rater variability, consistently high prediction bias, and high false discovery rate [2]. In fact, our end-to-end methodology demonstrates statistically significant improvement over PVA in false discovery rate, and non-significant improvement in assessment bias, the two key metrics reported in [2] (see Table 5). Furthermore, our proposed methodology has several intrinsic benefits relevant in the clinical setting. First, our methods can provide real-time inference, due to our single-pass inference approach. In addition, our automated method is intrinsically deterministic during inference and thus has zero intra-rater variance, allowing for more reliable stenosis estimation. While our end-to-end methodology still has room for improvement, its superior performance over PVA, as well as high inference speed and high intra-rater reliability make it well suited for adoption in the clinical setting.

Our current work focuses on still-frame image analysis of the RCA. One direction of future work would be to extend our approach to the other main coronary arteries, such as LAD and LCx. Here, the technical challenge is greater, as overlapping arterial segments adds additional challenges to the tasks of stenosis localization, segmentation, and classification. Additionally, our current work focused only on still-frame images. These still-frames are typically selected by a skilled QCA analyst to obtain the greatest image contrast at end-diastole with the least foreshortening and least amount of arterial overlap. Another direction of future work would be to identify and determine stenoses directly from multi-frame fluoroscopic imaging, including integrating imaging from multiple fluoroscopic views.

## 7. Conclusion

We establish a deep learning framework that characterizes the intensity of stenosis in still-frame invasive coronary angiography images of the right coronary artery. This framework involves a three-step process of localization, segmentation and classification. At each step, we design a CNN model that obtains high performance for the specific task. In addition, we develop an end-to-end deep learning method that integrates all three tasks in an automated pipeline, which demonstrates statistically significant performance improvement in false detection rate compared to physician visual assessment (PVA), the current standard for real-time coronary stenosis analysis in the clinical setting. This improvement over PVA, as well as our model's intrinsic high intra-rater reliability and real-time inference speed, make our approach well suited for adoption in the clinical setting. To our knowledge, this is the first time deep learning has been applied to replicate the QCA protocol, and the first time a fully-automated learning system has demonstrated significantly improved performance in analyzing coronary stenosis relative to PVA, the current clinical standard for real-time stenosis analysis.

# Appendix CNN Hyperparameter and Training Details

## A.1 General Training

- Epochs trained: 100
- Early stopping if no improvement in validation set across 20 epochs
- 5x learning rate reduction if no improvement in validation set across 5 epochs
- Batch size: 2
- Dataset split ratio 70:15:15 for train:validation:test
    - Stratified random sampling strata thresholds (based on stenosis degree):
        - 0%; 30%; 55%; 70%; 85%; 100%
- Dropout 0%
- Batch normalization used
- Data augmentation used during train phase

## A.2 Data Augmentation

- Random rotation: +/- 7 degrees
- Random shear: +/- 7 degrees
- Random shift: +/- 0.04 of image in both axes
- Elastic transformation: alpha=70; sigma=7 (ultimately unused)
- Uniform distribution used in all cases

## A.3 Localization CNN Details

- Input Dimension: 512x512
- Proposed CNN
    - Activation: Leaky RELU
    - Final activation: sigmoid
    - Final output dimension: $15 \times 15$
    - Loss: Weighted grid-based sigmoid cross entropy (see Equation 1)
    - All other hyperparameters based on implementation in [11]
- ResNet50 (Regression-based localization)
    - Feature aggregation layer: Global max pooling
    - Final activation: Linear
    - Final output dimension: 2
    - Loss function: Mean squared error
    - All other hyperparameters and model details used based on [8,30]

## A.4 Segmentation Details

- Input Dimension: 192x192 pixels
- Output Dimension: 192x192 pixels
- Morphologically-Weighted Cross Entropy Hyperparameters
    - $\alpha = 3$
    - $\beta = 64$
    - $\gamma = 128$
    - $\delta = 10$
    - $\sigma = 15$
- Proposed CNN
    - Pooling layer: average
    - Number of blocks: 5
    - Filters per block: 32/64/128/256/512
    - All other hyperparameters and model details used based on [13]
- DenseNet FCN
    - All hyperparameters and model details used based on [31]
- DeepLabV3 (Atrous convolutions)
    - CNN backbone: ResNet50
    - Upsampling type: Bilinear
    - All other hyperparameters and model details used based on [11,32]

## A.5 Classification Details

- Input Dimension: 192x192
- Evaluation via classification: Binary (+/- 70% stenosis)
- Training loss function: Mean squared error i.e. ordinal regression
- Proposed CNN Details
    - Model architecture (see Figure 5)
    - Final layer activation: Linear
- ResNet50
    - Aggregation layer: Global max pooling
    - Final activation layer: Linear
    - All other hyperparameters and model details used based on [8]

## A.6 End-to-End Details

- Pipeline consists of proposed CNN from each of three tasks integrated end to end (see Figure 1)
- Each model is pre-trained from individual tasks
- Final end-to-end has frozen layers for localization and segmentation modules.
- Pre-localized segmentation-classification model
    - Integrates segmentation and classification proposed CNNs
    - Segmentation and classification CNN weights pretrained from individual tasks
    - Segmentation module frozen in final model
    - Input is pre-localized 192x192 image input using localization oracle